\documentclass[lettersize,journal]{IEEEtran}
\usepackage{amsmath,amsfonts}
\usepackage{algorithmic}
\usepackage{array}
\usepackage[caption=false,font=normalsize,labelfont=sf,textfont=sf]{subfig}
\usepackage{textcomp}
\usepackage{stfloats}
\usepackage{url}
\usepackage{verbatim}
\usepackage{graphicx}
\usepackage{makecell}
\usepackage{textcomp}
\usepackage{stfloats}
\usepackage{url}
\usepackage{verbatim}
\usepackage{graphicx}
\usepackage{cite}
\usepackage{siunitx}  
\usepackage{textcomp} 
\usepackage{bm}    
\usepackage{tensor}
\usepackage{algorithm}

\usepackage{cite}

\makeatletter
\let\NAT@parse\undefined
\makeatother

\usepackage[colorlinks,linkcolor=blue,anchorcolor=black,citecolor=blue,urlcolor=blue,hyperfootnotes=true]{hyperref}   
\usepackage[all]{hypcap} 

\def\BibTeX{{\rm B\kern-.05em{\sc i\kern-.025em b}\kern-.08em
    T\kern-.1667em\lower.7ex\hbox{E}\kern-.125emX}}
\usepackage{balance}
\begin{document}

\title{Wearable Multimodal Human–Machine Interface for Integrated Hand Intentions Decoding in Dynamic Teleoperation}

\author{Jiaxuan Li$^{1\dag}$, Yinshi Wu$^{1\dag}$, Xiao Zhang$^{1}$, Hongyu Wang$^{1}$, Renzhen Le$^{1}$,\\ Zhenzhi Ying$^{2}$, and Liming Shu$^{1*}$, \IEEEmembership{Member,~IEEE}
\thanks{$^\dag$Equal contributions. $^*$Corresponding author.}
\thanks{$^{1}$Intelligent Equipment and Medical Device Laboratory, Department of Mechanical Engineering, Dalian University of Technology, Dalian 116024, China. 
        (email: jiaxuan@mail.dlut.edu.cn, l.shu@dlut.edu.cn)}%
\thanks{$^{2}$Manufacturing Laboratory, Department of Mechanical Engineering, The University of Tokyo, Tokyo 113-8656, Japan. 
       }}%


\maketitle

\begin{abstract}
Under ubiquitous teleoperation environments with optically challenging conditions, an interface for tele-operated grasping that combines wearability with precise decoding of hand intentions (hand pose, gestures, and grasping force) is essential. Yet, existing interfaces often fall short in meeting these demands, compromising either the diversity of multiple intentions decoding or wearability. To address this, we developed a novel Multiple Intentions Decoding Human–Machine Interface (MI-DHMI) that integrates high-throughput surface electromyography (sEMG) sensors with hand-mounted and forearm-mounted inertial measurement units (IMUs). The developed interface is supported by a unified framework for simultaneous multiple intentions decoding. By employing multimodal deep learning and hardware design with a low noise floor, the decoding framework selectively focuses on the sEMG components that are genuinely associated with finger movements. This effectively reduces decoding errors caused by sEMG variability during unconstrained upper-limb motions, thereby significantly enhancing robustness. Even under unconstrained wrist and forearm motion, the interface achieves a gesture recognition accuracy exceeding 97\%, grasping force estimation with $R^2 = 0.95$, and hand pose decoding consistent with the actual hand pose, outperforming baseline devices and algorithms. Ablation studies further validate the effectiveness of the proposed decoding framework. Finally, two online experiments were conducted to validate the device, demonstrating its superior performance in high-stability tasks, including a pouring task and object grasping. The developed interface provides a new solution of a fully wearable, multiple intentions decoding system, offering effective support for ubiquitous teleoperation and contributing to the advancement of human--machine interaction research.
\end{abstract}


\section{Introduction}
\IEEEPARstart{R}{emote} robot teleoperation has become an essential paradigm in high-risk domains, such as operations at height and the handling of hazardous biological or chemical materials~\cite{10035484}. In these scenarios, operators are required to perform fine manipulation and grasping tasks remotely in dynamic and unstructured environments without reliance on fixed workspaces, while maintaining strict safety and reliability. This demands human–machine interfaces (HMIs) that can accurately capture and transmit comprehensive hand intentions (hand pose, gestures, and grasping force) under unrestricted wrist and forearm motion, while being highly portable and fully wearable.

Existing teleoperation interfaces often fail to simultaneously achieve multiple intentions decoding and rapid workspace transfer. Indirect controllers, such as joysticks or master manipulators~\cite{{anipulator-based}}, provide reliable, low-latency control but cannot decode hand gestures and their heavy structures limit fast workspace transitions. Optical-based systems~\cite{li2022dexterous,huang2025human,li2025six,RGBhandover,li2020mobile} capture detailed kinematics but rely on external infrastructure, are sensitive to occlusions, lighting, and viewpoints, and generally cannot capture hand force. Wearable gloves~\cite{lu2025ultra,liu2024reconfigurable,yu2024compact,belcamino2024systematic} facilitate more natural interaction but often constrain hand mobility and may compromise the user’s native tactile perception. As a result, existing interfaces face inherent trade-offs between the diversity of decoded hand intentions and usability, particularly in anytime–anywhere teleoperation scenarios involving unrestricted human motion.

To address these limitations, recent studies have explored the feasibility of integrating wearable devices into teleoperation interfaces. Inertial measurement units (IMUs), as wearable sensors, can decode hand gestures and poses without relying on optical information~\cite{lu2023measurement,imu_gesture}. However, IMUs alone cannot provide sufficient information for decoding grasping forces. Consequently, physiological signals related to muscle force have been introduced into HMIs. Common biosignals, such as surface electromyography (sEMG)~\cite{kaifosh2025generic,sEMG_gesture_force,liu2024human,meng2025real,kusuru2024improved} and pressure-based force myography (pFMG)~\cite{sEMG_pFMG_8834,10177680}, have been applied to decode gesture and force. By fusing sEMG and IMU signals, the complementary information of muscle activation and hand kinematics can be jointly exploited, enabling richer and more reliable decoding of hand intentions~\cite{li2024improving, mao2023simultaneous, zhang2026comprehensive}.
\IEEEpubidadjcol

\begin{figure*}[!t]
\centering
\includegraphics[width=\textwidth]{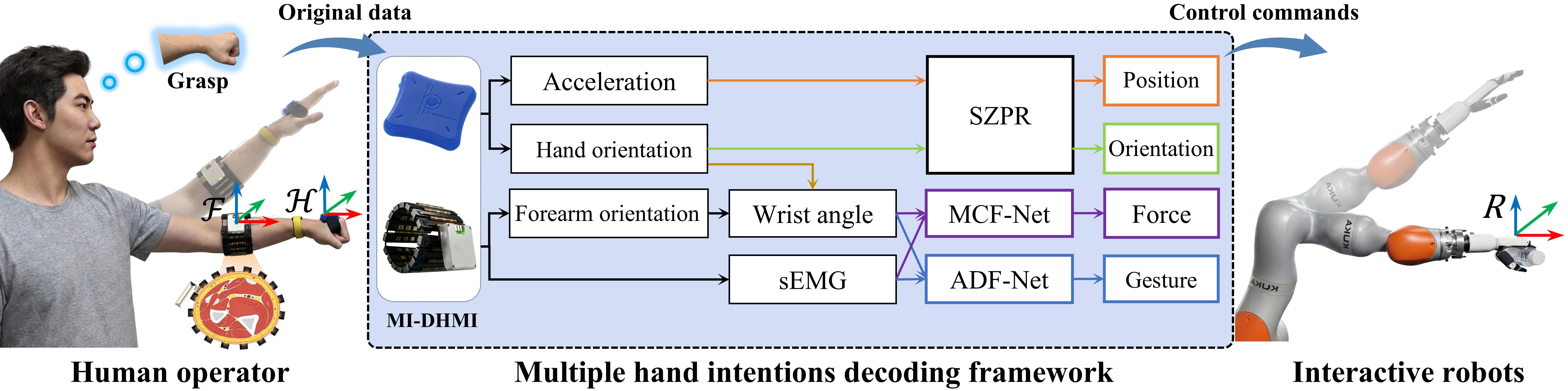}
\caption{Schematic design overview of the proposed MI-DHMI for multiple informations decoding in dynamic teleoperation.}
\label{fig_1}
\end{figure*}

Although sEMG–IMU fusion has proven effective for gesture and sign language recognition~\cite{wang2026fusion}, its use in teleoperation interfaces remains limited. In practical teleoperation settings, hand gestures rarely occur in isolation but are accompanied by wrist rotation and forearm motion. Such coupled movements introduce pronounced variability in sEMG signals~\cite{chen2025noise} and substantially reduce the reliability of gesture and force estimation. As a result, stable decoding of hand gesture intentions in realistic teleoperation scenarios, while preserving a fully wearable interface, remains an open challenge. Motivated by this gap, we developed a wearable multimodal HMI that integrates a unified hardware system with a multiple intentions decoding framework, enabling robust intentions decoding in dynamic teleoperation scenarios, as illustrated in Fig.~\ref{fig_1}.

The main contributions of this paper are as follows. 

1) We developed a wearable Multiple Intentions Decoding HMI (MI-DHMI) that integrates high-throughput sEMG acquisition with IMU sensing on the hand and forearm, enabling the acquisition of hand–forearm kinematics and sEMG signals while preserving wearability.

2) We proposed a unified decoding framework capable of simultaneously decoding multiple hand intentions under unconstrained wrist and forearm motion. The framework integrates an Attention-based~\cite{vaswani2017attention,10086669} Dual-level Fusion Network (ADF-Net), a Motion-Compensated Force Network (MCF-Net), and a Soft Zero-Velocity Update (Soft ZUPT)-based Pose Reconstruction (SZPR) module, enabling simultaneous inference of hand gestures, force, and poses within a single cohesive architecture.

3) Through ablation studies and online teleoperation experiments, we validated the effectiveness of the proposed interface, demonstrated the advantages of multimodal fusion, and showed superior decoding performance in dynamic teleoperation scenarios.

To the best of our knowledge, this work is the first fully wearable multimodal HMI for ubiquitous teleoperated grasping tasks that simultaneously decodes hand intentions (hand poses, gestures, and grasping force) under unconstrained wrist and forearm motion.

The wearable MI-DHMI and the decoding framework are introduced in Sec.~\ref{sec:II}, and the experimental protocol is presented in Sec.~\ref{sec:III}. We then present the experimental results in Sec.~\ref{sec:IV}, followed by the discussion in Sec.~\ref{sec:V}, and the conclusion in Sec.~\ref{sec:VI}.

\section{System}
\label{sec:II}

\subsection{System Overview}
\begin{figure}[!t]  
\centering
\includegraphics[width=\columnwidth]{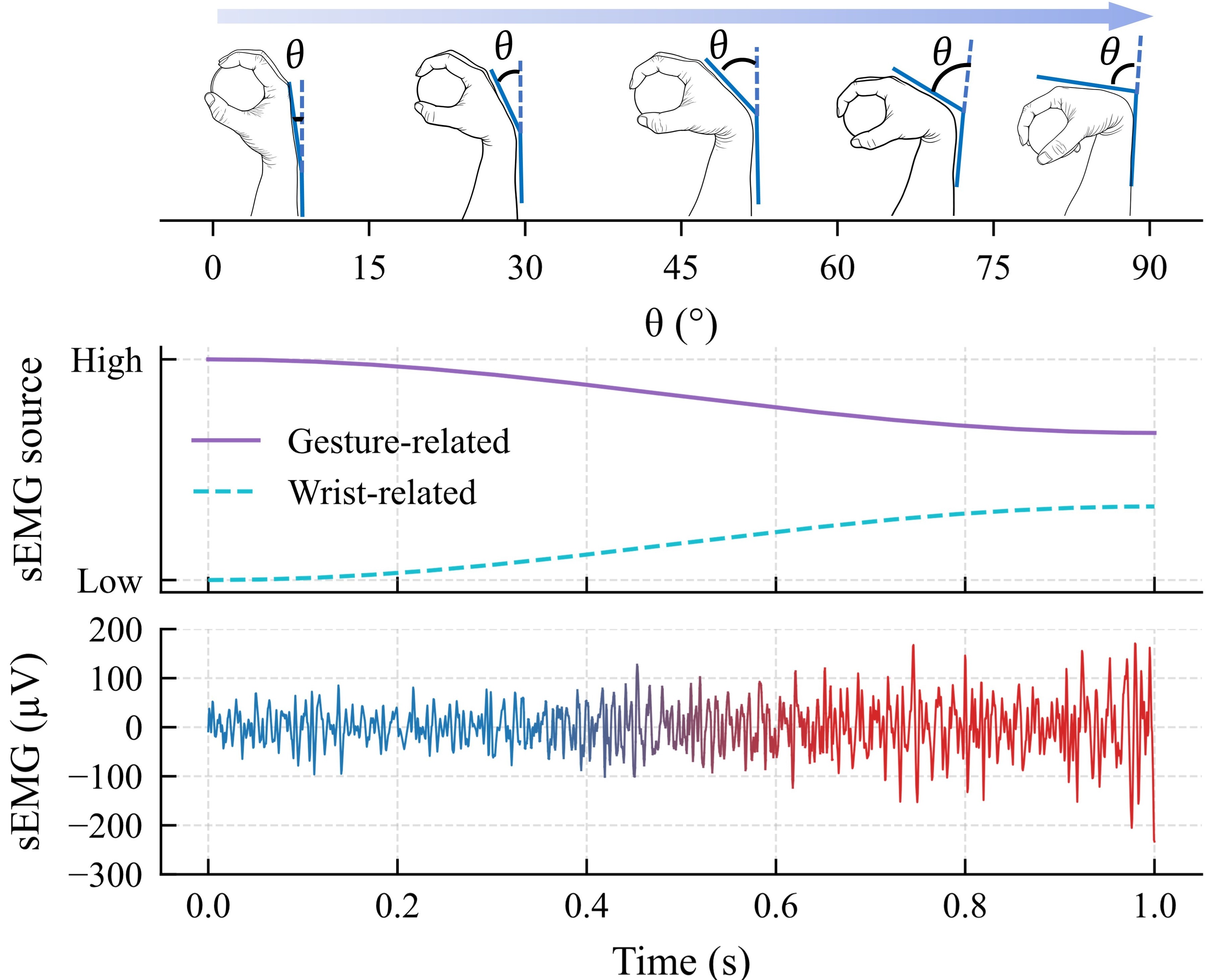}
\caption{Illustration of sEMG signal coupling caused by simultaneous wrist rotation and hand gesture execution, which degrades gesture intention and grasping force decoding performance.}
\label{fig_3}
\end{figure}

In this section, we introduce the wearable MI-DHMI developed in this work and its associated decoding framework, as shown in Fig.~\ref{fig_1}. The MI-DHMI comprises a forearm armband and a hand-mounted IMU. The forearm armband integrates a 64-channel sEMG sensor and an IMU to acquire forearm sEMG signals and orientation data, while the hand-mounted IMU captures acceleration and orientation information. All data are transmitted wirelessly to a central computing unit via a local area network. Hand pose is reconstructed based on measurements from the hand-mounted IMU. Wrist rotation introduces substantial variability into sEMG signals, which can interfere with gesture-related features and force prediction, as illustrated in Fig.~\ref{fig_3}. To address this issue, wrist angles obtained by combining orientation information from both the hand and forearm IMUs are fused with the sEMG data, and this enables robust decoding of hand gestures and grasping forces under unconstrained wrist and forearm motion. The following subsections provide a detailed description of each module within the proposed system.

\subsection{Notation and Coordinate Systems}
Scalars are denoted by non-bold symbols $x, X \in \mathbb{R}$, vectors by bold lowercase symbols $\boldsymbol{x} \in \mathbb{R}^n$, and matrices by bold uppercase symbols $\boldsymbol{X} \in \mathbb{R}^{n \times m}$. The estimated values are indicated by a hat $(\hat{\cdot})$. The time-varying vectors are denoted as $\boldsymbol{p}(t)$. 

A vector expressed in the world frame $\left\{\mathcal{W}\right\}$ is denoted as ${}^{W}\boldsymbol{p}$. The rotation from frame $\left\{\mathcal{R}\right\}$ to frame $\left\{\mathcal{W}\right\}$ is represented by the quaternion ${}^{W}_{R}\boldsymbol{q} = [q_w, q_x, q_y, q_z]^{\mathrm{T}}$. The operator $\otimes$ denotes quaternion multiplication. 

An inertial world frame $\{\mathcal{W}\} = \{^{W}\boldsymbol{p},\,^{W}\boldsymbol{x},\,^{W}\boldsymbol{y},\,^{W}\boldsymbol{z}\}$ is defined, with its $^{W}\boldsymbol{z}$-axis opposite to the direction of gravity. The coordinate frames of individual system components are introduced in their respective subsections. For simplicity, when a vector is expressed in the world frame, the superscript is omitted, i.e., $\boldsymbol{p}$ is used instead of ${}^{W}\boldsymbol{p}$.
The target position and orientation commands sent to the controlled device are denoted as $\boldsymbol{p}_{R_c}$ and ${}^{W}_{R}\boldsymbol{q}_c$, respectively.

\subsection{Hardware Architecture of the MI-DHMI}

\begin{figure}[!t]  
\centering
\includegraphics[width=\columnwidth]{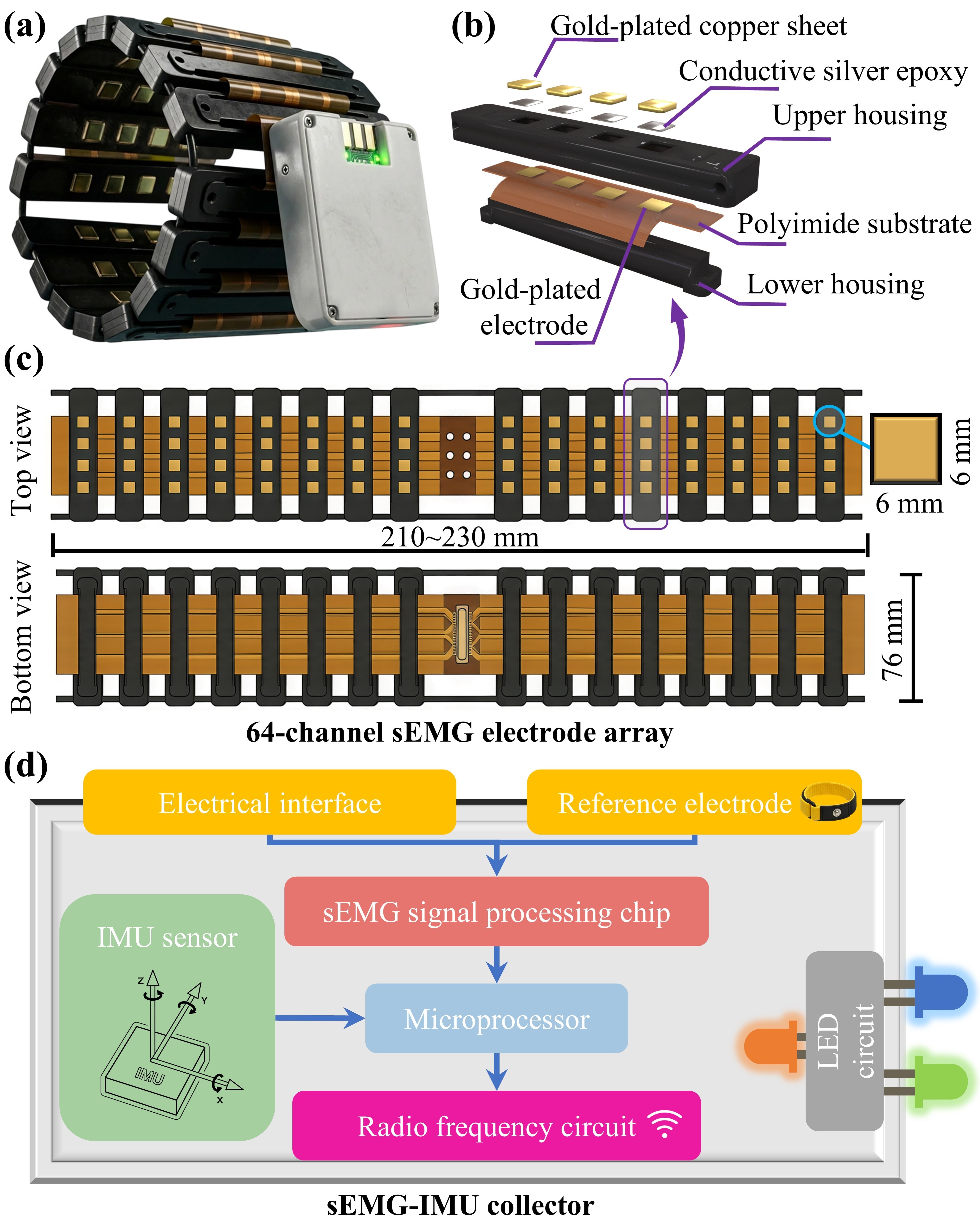}
\caption{(a) The stretchable armband in the HI-DHMI, featuring a 64-channel sEMG electrode array and an sEMG–IMU collector. (b) Schematic exploded view showing the hybrid rigid–flex design of a single block in the stretchable armband. Each block contains four electrodes, encapsulated in a 3D-printed housing. (c) Schematic diagram of the 64-channel sEMG electrode array. (d) Internal view of the sEMG–IMU collector.}
\label{fig_4}
\end{figure}

The MI-DHMI consists of an armband and a hand-mounted IMU. The armband integrates a stretchable 64-channel electrode array and an sEMG-IMU acquisition module, as shown in Fig.~\ref{fig_4}. The 64-channel gold-plated copper dry electrode array is uniformly arranged in a 4×16 grid, providing a total of 64 acquisition channels. The electrodes are spaced 12~mm apart along the forearm axis. The armband is available in three circumferential sizes (210, 220, and 230~mm) and can be adjusted to fit different forearm sizes using elastic cords. The electrode array is fabricated on a flexible polyimide substrate. The sEMG signals are sampled at \SI{2}{\kilo\hertz} with a noise level of \SI{3.14}{\micro\volt_{\mathrm{rms}}}, while the IMU in the sEMG-IMU acquisition module collects acceleration and orientation data at \SI{400}{\hertz}. The integrated IMU measures arm kinematics at \SI{400}{\hertz}. Hand kinematics are measured using an LPMS-B2 IMU (Alubi Inc., China). This IMU operates at \SI{400}{\hertz} and weighs \SI{12}{\gram}. The total weight of the MI-DHMI is approximately \SI{161}{\gram}.

As shown in Fig.~\ref{fig_1}, we define the forearm frame
$\{\mathcal{F}\} = \{^{F}\boldsymbol{p},\,^{F}\boldsymbol{x},\,^{F}\boldsymbol{y},\,^{F}\boldsymbol{z}\}$
and the hand frame
$\{\mathcal{H}\} = \{^{H}\boldsymbol{p},\,^{H}\boldsymbol{x},\,^{H}\boldsymbol{y},\,^{H}\boldsymbol{z}\}$.
The hand IMU is mounted at the center of the dorsum. The $^{H}\boldsymbol{x}$ axis is aligned with the middle finger, and the $^{H}\boldsymbol{z}$ axis is perpendicular to the dorsum.
For the forearm IMU, the $^{F}\boldsymbol{x}$ axis follows the longitudinal direction of the forearm, and the $^{F}\boldsymbol{z}$ axis is normal to the forearm surface at the mounting location.
The $^{H}\boldsymbol{y}$ and $^{F}\boldsymbol{y}$ axes are defined by the right-hand rule.

\subsection{Hand Pose Estimation in the Decoding Framework}


Within the proposed decoding framework, hand pose estimation is performed using the SZPR method, which represents the hand pose in terms of both position and orientation. To compensate for the fixed misalignment between the IMU world frame and the world frame used in this work, a constant rotational offset is applied:
\begin{equation}
{}^{W}_{H}\hat{\boldsymbol{q}}(t)
 = 
\boldsymbol{q}_{\mathrm{offset}}
\otimes
\boldsymbol{q}_{\mathrm{IMU}}(t),
\end{equation}
where $\boldsymbol{q}_{\mathrm{offset}}$ is a constant calibration quaternion. For position estimation, the overall procedure is summarized in Algorithm~\ref{algo:imu_soft_zupt_simplified}, and the detailed steps are outlined below.

The IMU provides tri-axial linear acceleration measurements $^{H}\boldsymbol{a}(t)$ in the hand frame, which are transformed into the world frame via the rotation matrix:
\begin{equation}
\scalebox{0.858}{$
\boldsymbol{R(q)} =
\begin{bmatrix}
1 - 2(q_y^2 + q_z^2) & 2(q_xq_y - q_zq_w) & 2(q_xq_z + q_yq_w) \\
2(q_xq_y + q_zq_w) & 1 - 2(q_x^2 + q_z^2) & 2(q_yq_z - q_xq_w) \\
2(q_xq_z - q_yq_w) & 2(q_yq_z + q_xq_w) & 1 - 2(q_x^2 + q_y^2)
\end{bmatrix},
$}
\end{equation} where $\|\boldsymbol{q}\| = 1$. The world frame acceleration is then obtained as:
\begin{equation}
\hat{\boldsymbol{a}}_{H}(t)
= \boldsymbol{R}^\top\bigl({}^{W}_{H}\hat{\boldsymbol{q}}(t)\bigr)
\, {}^{H}\boldsymbol{a}(t).
\end{equation}

Velocity and position are obtained through discrete-time integration of the world-frame acceleration. To mitigate drift caused by sensor noise and bias, we employ a Soft ZUPT strategy. Specifically, a time-varying Soft ZUPT gain $\alpha(t) \in [0,1]$, is maintained and updated based on a motion state indicator $\mathit{s}(t)$. This gain is used to mitigate velocity drift. When the hand is detected as stationary ($s(t)=1$), $\alpha(t)$ increases to slow down the velocity estimate. Conversely, when the hand is moving ($s(t)=0$), $\alpha(t)$ decreases to allow the velocity to follow the actual motion. 
The velocity is updated as:
\begin{equation}
\alpha(t)
=
\mathrm{clip}
\Bigl(
\alpha(t-1)
+
\delta\, s(t)
-
\gamma\, \bigl(1 - s(t)\bigr),
\; 0,\; 1
\Bigr),
\end{equation}
\begin{equation}
{\hat{\boldsymbol{\mathit{v}}}_{H}(t)} =
\bigl(1 - \beta \alpha(t)\bigr)
\left[
{\hat{\boldsymbol{\mathit{v}}}_{H}(t-1)}
+ d_t \, {}\hat{\boldsymbol{a}}_{{H}}(t)
\right].
\end{equation}
where $\delta$ denotes the gain increment when the sensor is stationary, $\gamma$ denotes the gain decrement when the sensor is moving, and $\mathit{s}(t)$ indicates whether the system is in a stationary state. The parameter $d_t$ corresponds to the IMU sampling interval, while $\beta$ is a scaling factor that regulates the attenuation strength.

\begin{algorithm}[bt]
\caption{Hand Position Estimation with Soft ZUPT}
\label{algo:imu_soft_zupt_simplified}
\begin{algorithmic}[1]

\REQUIRE
1) ${}^{H}\boldsymbol{a}(t) \in \mathbb{R}^3$, tri-axial IMU linear acceleration expressed in the hand frame $\{\mathcal{H}\}$; \\
2) ${}^{H}\boldsymbol{\omega}(t) \in \mathbb{R}^3$, tri-axial IMU angular velocity expressed in the hand frame; \\
3) ${}^{W}_{H}\hat{\boldsymbol{q}}(t) \in \mathbb{H}$, estimated unit quaternion representing the IMU orientation with respect to the world frame; \\
4) $d_t$, IMU sampling interval; \\
5) ${}\boldsymbol{v}_{H}(t_0) \in \mathbb{R}^3$, initial hand velocity in the world frame; \\
6) ${}\boldsymbol{p}_{H}(t_0) \in \mathbb{R}^3$, initial hand position in the world frame; \\
7) $\alpha(t_0) \in [0,1]$, initial Soft ZUPT gain; \\
8) $s(t_0) \in \{0,1\}$, initial motion state indicator.

\ENSURE
${}\hat{\boldsymbol{p}}_{H}(t) \in \mathbb{R}^3$, estimated hand position.

\STATE Initialize variables:
$t = t_0$,
${}\boldsymbol{v}_{H}(t_0) = \boldsymbol{0}$,
${}\boldsymbol{p}_{H}(t_0) = \boldsymbol{0}$,
$\alpha(t_0) = 0.4$,
$s(t_0) = 0$
\STATE Set parameters: $\delta = 0.05$, $\gamma = 0.1$, $\beta = 0.6$
\WHILE{system is running}
    \STATE ${}{{t}}
    \gets
    {}{{t}}+1$
    \STATE ${}\hat{\boldsymbol{a}}_{H}(t)
    \gets
    \boldsymbol{R}\!\left({}^{W}_{H}\hat{\boldsymbol{q}}(t)\right)
    \, {}^{H}\boldsymbol{a}(t)$

    \STATE $s(t) \gets 
    \arg\max \Bigl(
    f_{\mathrm{MLP}}\Bigl(
    \bigl[ {}^{H}\boldsymbol{a}(t-k),\, {}^{H}\boldsymbol{\omega}(t-k) \bigr]_{k=0}^{5}
    \Bigr)
    \Bigr)$
    
    \STATE $\alpha(t) \gets
    \mathrm{clip}\!\Bigl(
    \alpha(t-1)
    +
    \delta\, s(t)
    -
    \gamma\, \bigl(1 - s(t)\bigr),
    \; 0,\; 1
    \Bigr)$
    
    \STATE ${}\hat{\boldsymbol{v}}_{H}(t)
    \gets
    \left(1 - \beta \alpha(t)\right)
    \left[
    {}\hat{\boldsymbol{v}}_{H}(t-1)
    +
    d_t \, {}\hat{\boldsymbol{a}}_{H}(t)
    \right]$

    \STATE ${}\hat{\boldsymbol{p}}_{H}(t)
    \gets
    {}\hat{\boldsymbol{p}}_{H}(t-1)
    +
    d_t \, {}\hat{\boldsymbol{v}}_{H}(t)$

\ENDWHILE

\end{algorithmic}
\end{algorithm}

The motion state indicator $\mathit{s}(t)$ is estimated using a lightweight multilayer perceptron (MLP). A sliding-window strategy is adopted to generate input samples $\boldsymbol{x}(t)$ for the MLP. At each time step $t$, the tri-axial linear acceleration $\boldsymbol{a}(t) \in \mathbb{R}^3$ and angular velocity $\boldsymbol{\omega}(t) \in \mathbb{R}^3$, along with their measurements from the previous five sampling instants, are stacked to form the input feature vector. The MLP model consists of three hidden layers with 64, 128, and 64 neurons, respectively, followed by a two-dimensional output layer. The motion state indicator $\mathit{s}(t)$ is obtained by applying an argmax operation to the network output:
\begin{equation}
\mathit{s}(t) = \arg\max \Bigl(
    f_{\mathrm{MLP}}\Bigl(
    \bigl[ {}^{H}\boldsymbol{a}(t-k),\, {}^{H}\boldsymbol{\omega}(t-k) \bigr]_{k=0}^{5}
    \Bigr)
    \Bigr).
\end{equation}
The indicator $\mathit{s}(t)$ is subsequently used to adaptively regulate the Soft ZUPT gain. Subsequently, the hand position in the world coordinate system is calculated by performing a forward euler integral on the decayed velocity.


\subsection{Gesture Recognition in the Decoding Framework}

The hand gesture recognition module decodes gesture intentions mainly from sEMG signals. To compensate for sEMG disturbances induced by unconstrained wrist and forearm motions, IMU signals are fused with sEMG, where hand IMU quaternions are expressed in the forearm IMU frame to represent wrist angles. The hand orientation relative to the forearm IMU is computed via an initialization-based quaternion calibration:
\begin{equation}
{}^{F}_{H}\hat{\boldsymbol{q}}(t)
=
{}^{W}_{F}\hat{\boldsymbol{q}}(t_0)
\otimes
\bigl({}^{W}_{H}\hat{\boldsymbol{q}}(t_0)\bigr)^{-1}
\otimes
{}^{W}_{H}\hat{\boldsymbol{q}}(t),
\end{equation}
where $t_0$ denotes the initialization instant with the wrist held in a neutral, stationary pose.

\begin{figure}[!t]  
\centering
\includegraphics[width=\columnwidth]{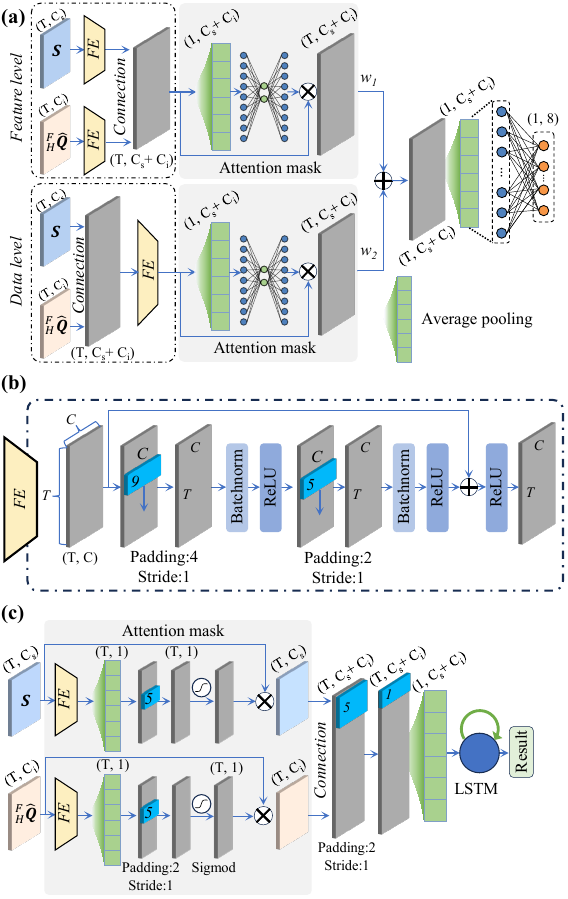}
\caption{(a) Architecture of the proposed ADF-Net for gesture intention recognition, featuring dual-branch data-level and feature-level fusion of sEMG and IMU signals, an FE module, and a lightweight attention mask module for enhanced feature representation. (b) Internal architecture of the FE module. (c) Network architecture of the proposed MCF-Net, which fuses sEMG and hand–forearm quaternion features using temporal attention to mitigate wrist-motion-induced sEMG fluctuations for robust force regression.}
\label{fig_5}
\end{figure}

We propose an ADF-Net for gesture recognition as shown in Fig.~\ref{fig_5}(a), which integrates complementary information from sEMG and IMU signals through both data-level and feature-level fusion. Feature extraction is automatically learned via deep learning, avoiding manual feature engineering. A one-dimensional convolutional neural network (1D-CNN) is adopted as the feature extraction (FE) backbone, where kernels slide along the temporal dimension and jointly process all channels as shown in Fig.~\ref{fig_5}(b). Residual connections are incorporated to improve generalization while preserving the input dimensionality. 

\subsubsection{Data-level fusion} 
The sEMG time-window matrix $\boldsymbol{S}\in\mathbb{R}^{T\times C_s}$ ($C_s=64$) and the hand--forearm
quaternion matrix ${}^{F}_{H}\hat{\boldsymbol{Q}}\in\mathbb{R}^{T\times C_i}$ ($C_i=4$) are concatenated
along the channel dimension. To reduce inter-modality magnitude discrepancies, sEMG signals are
normalized to $[0,1]$ on a per-channel basis. The fused input is
$\boldsymbol{X} = [\boldsymbol{S}, {}^{F}_{H}\hat{\boldsymbol{Q}}] \in \mathbb{R}^{T \times (C_s + C_i)}$, which is fed into the FE module. The extracted features are given by
\begin{equation}
\boldsymbol{F}_d = f_{\mathrm{FE}}(\boldsymbol{X}) \in \mathbb{R}^{T \times (C_s + C_i)},
\end{equation}
where both the temporal length and channel dimensionality are preserved, enabling the network to learn coupled representations of sEMG activity and wrist kinematics.

\subsubsection{Feature-level fusion} 
Features are first extracted independently from each modality:
\begin{equation}
\boldsymbol{F}_s = f_{\mathrm{FE}}(\boldsymbol{S}) \in \mathbb{R}^{T \times C_s}, \quad 
\boldsymbol{F}_i = f_{\mathrm{FE}}({}^{F}_{H}\hat{\boldsymbol{Q}}) \in \mathbb{R}^{T \times C_i},
\end{equation}
The extracted features are then concatenated as $\boldsymbol{F}_f=[\boldsymbol{F}_s,\boldsymbol{F}_i]\in\mathbb{R}^{T \times (C_s + C_i)}$, which is subsequently used for classification. This fusion mitigates insufficient feature capture due to modality heterogeneity and improves decoding robustness. 

\subsubsection{Attention mask module} 
We introduce a lightweight attention module to emphasize the most informative features. An average pooling layer first compresses the temporal dimension into a fixed-length feature vector. This vector is then processed by a single-layer encoder–decoder, which maps it to a low-dimensional latent space and reconstructs it to generate a channel-wise attention weight vector $\mathbf{a}_c\in \mathbb{R}^{C}$. The weight vector is applied along the feature dimension to the input of the attention mask module.
The output of the attention mask module is:
\begin{equation}
\tilde{\boldsymbol{F}} = \boldsymbol{F} \odot \mathbf{a}_c,
\end{equation}
where $\boldsymbol{F}$ denotes the input of the attention mask module (either $\boldsymbol{F}_f$ or $\boldsymbol{F}_d$), $\odot$ denotes element-wise multiplication along the feature dimension, and $\tilde{\boldsymbol{F}}$ is the weighted output.

\subsubsection{Classification module}

Finally, the outputs of the data-level and feature-level branches are combined via a weighted summation, where the weights are learned during training. The fused feature vector is aggregated through temporal average pooling and then fed into a two-layer fully connected network for classification. This dual-branch design allows the model to exploit complementary information from both fusion strategies, ensuring robustness and enhancing discriminative performance.

\subsection{Force Regression in the Decoding Framework}

In the framework, the force regression module employs a novel algorithm, termed MCF-Net, which integrates time-windowed sEMG samples $\boldsymbol{S}$ with hand–forearm quaternion windows ${}^{F}_{H}\hat{\boldsymbol{Q}}$ to compensate for sEMG fluctuations induced by unconstrained wrist motion, as shown in Fig.~\ref{fig_5},(c).

Each modality is first processed independently by a FE module to obtain high-level temporal representations:
\begin{equation}
\boldsymbol{F}_s
=
f_{\mathrm{FE}}(\boldsymbol{S})
\in
\mathbb{R}^{T \times C_s},
\quad
\boldsymbol{F}_i
=
f_{\mathrm{FE}}({}^{F}_{H}\hat{\boldsymbol{Q}})
\in
\mathbb{R}^{T \times C_i},
\end{equation}
where $\boldsymbol{F}_s$ and $\boldsymbol{F}_i$ denote the extracted features of the sEMG and quaternion modalities, respectively. 

To enhance temporal representation, a lightweight temporal attention mechanism is applied independently to each modality.
For a given modality $m \in \{s, i\}$, average pooling is first applied along the feature dimension to compress $\boldsymbol{F}_m$ into a temporal descriptor.
This descriptor is then processed by a 1D convolutional layer to capture local temporal dependencies and expand the temporal receptive field, followed by a sigmoid activation to generate a modality-specific temporal attention vector $\mathbf{a}_t^{m} \in \mathbb{R}^{T}$.

The temporal attention vector is applied to the corresponding feature representation via element-wise multiplication along the temporal dimension, combined with a residual connection:
\begin{equation}
\tilde{\boldsymbol{F}}_{m}
=
\boldsymbol{F}_{m}
\odot
\mathbf{a}_t^{m}
+
\boldsymbol{F}_{m},
\quad
m \in \{s, i\},
\end{equation}
where $\odot$ denotes element-wise multiplication. This mechanism allows the
network to emphasize force-relevant temporal segments while suppressing
interference caused by wrist rotation.

The temporally enhanced features from the two modalities are then concatenated
along the feature dimension and processed by a two-layer 1D convolutional
network for deep feature integration:
\begin{equation}
\boldsymbol{F}_{\text{fused}}
=
\mathrm{Conv1D}
\bigl(
[\tilde{\boldsymbol{F}}_s, \tilde{\boldsymbol{F}}_i]
\bigr)
\in
\mathbb{R}^{T \times (C_s + C_i)}.
\end{equation}

Subsequently, temporal average pooling is applied to extract the average representation of the current time window, which is then input to a Long Short-Term Memory network to update its hidden states and estimate the current grasp force. This enables each prediction to capture both the temporal dynamics of the current window and the sequential dependencies from prior predictions.

\section{Experiment}
\label{sec:III}
\subsection{Data Acquisition}

\begin{figure}[!t]  
\centering
\includegraphics[width=\columnwidth]{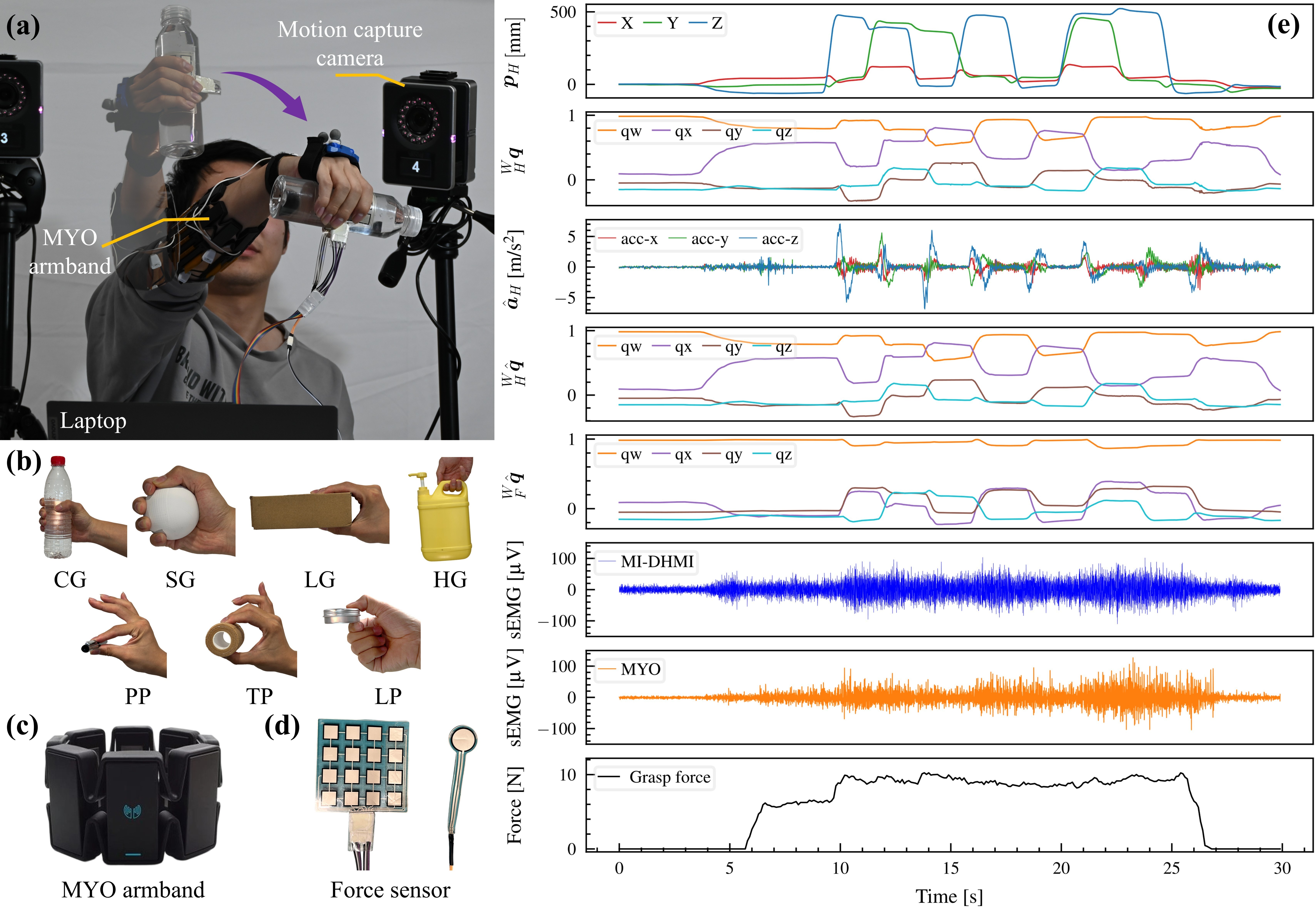}
\caption{(a) Experimental diagram. 
(b) Seven grasping gestures used in the experiments, including cylindrical, spherical, lumbrical, hook, precision pinch, tripod pinch, lateral pinch, and a rest gesture. 
(c) Schematic diagram of the MYO armband. 
(d) Schematic diagram of the force sensor. 
(e) Each row of subplots presents: 
(1) ground-truth human hand positions captured by the motion capture camera, 
(2) human hand orientations captured by the motion capture camera, 
(3) acceleration estimated by the hand-mounted IMU, 
(4) human hand orientations estimated using the hand-mounted IMU, 
(5) human forearm orientations estimated using the MI-DHMI, 
(6) sEMG signals collected by the MI-DHMI, 
(7) sEMG signals collected by the MYO armband, and 
(8) forces measured by the force sensor.}
\label{fig_6}
\end{figure}

As shown in Fig.~\ref{fig_6}, we simultaneously recorded sEMG signals, grasping force, and IMUs data during grasping. In addition, the motion capture system (Vicon, UK) was used to capture the ground-truth spatial hand poses, as illustrated in Fig.~\ref{fig_6}(a). Seven functional grasping gestures were selected based on~\cite{sharma2024automated}, considering their frequent use in daily activities, as shown in Fig.~\ref{fig_6}(b). Ten participants performed these gestures, including cylindrical (CG), spherical (SG), lumbrical (LG), hook (HG), precision pinch (PP), tripod pinch (TP), lateral pinch (LP), and a rest gesture (Re).  Prior to data collection, written informed consent was obtained from each participant. The experimental protocol was approved by the Ethics Committee of Dalian University of Technology (approval number: DUTSME230406-01) and was conducted in accordance with the Declaration of Helsinki. Each gesture was held for at least 15 seconds and five sessions were conducted per participant. Participants were allowed to move their arms and rotate their wrists freely, ensuring that the recorded sEMG captured both finger and wrist activities. MYO armband (Thalmic Labs Inc., Canada) were worn simultaneously to collect synchronous data for comparison with the proposed MI-DHMI.

For data acquisition, the MYO armband recorded sEMG signals at 200 Hz and IMU signals at 50 Hz, while force data were sampled at 10 Hz via a force sensor. The sEMG sampling rate served as the temporal reference for synchronizing all devices. For the MI-DHMI, sEMG from the armband, together with IMU signals from the armband and hand, as well as force data, were resampled to 2000 Hz. For the MYO armband, sEMG, IMU, and force signals were resampled to 200 Hz to align with the sEMG temporal resolution. Both sEMG and IMU signals were segmented using a sliding window of 75 ms with a step size of 37.5 ms, and the force label for each window was defined as the average force within the corresponding window.

Data were split on a session basis into training, validation, and test sets with a ratio of 3:1:1 to prevent data leakage. Both gesture classification and force regression networks were trained for 50 epochs on an NVIDIA GeForce RTX 3060 Ti GPU using Python 3.10.13 and PyTorch 2.5.1, with cross-entropy loss for gestures and mean squared error loss for force. The force regression models were trained separately for each individual grasping gesture.

\subsection{Ablation and Comparative Experiments}

Ablation and comparative experiments were conducted for both gesture recognition and force regression tasks using the proposed MI-DHMI. Its performance was compared with that of the MYO armband. Modal ablation experiments were performed to quantify the contribution of each modality to overall decoding performance. In addition, the attention modules in ADF-Net and MCF-Net were ablated separately to assess their respective impact on gesture recognition and force regression performance.

For gesture recognition, ADF-Net was compared with several classical machine learning models as baselines, including Decision Tree (DT), Random Forest (RF), K-Nearest Neighbors (KNN), and MLP. As these models rely on handcrafted features, the mean absolute value (MAV) was extracted from each sEMG channel as input features. For an sEMG segment $\boldsymbol{S} \in \mathbb{R}^{T \times C_s}$, the MAV of the $c$-th channel is defined as:
\begin{equation}
\mathrm{mav}_c = \frac{1}{T} \sum_{t=1}^{T} \left| \boldsymbol{S}_{c,t} \right|.
\end{equation}

A systematic comparison of data-level, feature-level, and hybrid fusion strategies was conducted to determine the optimal multimodal fusion approach for gesture decoding. Classification performance was evaluated using gesture recognition accuracy, while force regression performance was quantified by the coefficient of determination ($R^2$).

\subsection{Teleoperation Experiments}

To evaluate the practicality and robustness of the proposed MI-DHMI in dynamic teleoperation scenarios, we designed two teleoperation tasks to verify the feasibility of the entire framework. The first task consisted of the following sequential stages:
(i) guiding the manipulator end-effector toward a water bottle suspended in midair and grasping it with the robotic hand,
(ii) controlling the rotation of the manipulator end-effector to pour water into a bucket placed on a table, and
(iii) placing the empty bottle onto a designated target tray.
The second task consisted of the following sequential stages:
(i) guiding the robotic arm end effector to approach an object in a suitable orientation and performing a specific action to grasp the object,
(ii) controlling the robotic arm end effector to move to a designated location and placing the object inside a designated box.

During the task, a coordinate frame $\left\{\mathcal{R}\right\}$ was assigned to the robotic hand mounted at the manipulator end-effector, analogous to the human hand frame $\left\{\mathcal{H}\right\}$. The robotic hand's position is updated by adding the mapped hand position increment:
\begin{equation}
\begin{aligned}
{} \boldsymbol{p}_{R_c}(t+1)
&=
{} \boldsymbol{p}_{R}(t)
+
\Delta {} \boldsymbol{\hat{p}}_{H}(t), \\
\Delta {} \boldsymbol{\hat{p}}_{H}(t)
&=
\boldsymbol{\hat{p}}_{H}(t)
-
\boldsymbol{\hat{p}}_{H}(t-1).
\end{aligned}
\end{equation}
Similarly, the orientation of the human hand was directly mapped to update the orientation of robotic hand:
\begin{equation}
{}^{W}_{R}{\boldsymbol{q}}_c(t+1)
=
{}^{W}_{H}\hat{\boldsymbol{q}}(t).
\end{equation}

The ADF-Net decoded hand gestures into discrete grasping and releasing commands for the robotic hand. Grasping force was estimated with MCF-Net after each gesture transition, and maintained when gestures remained unchanged to ensure stable manipulation. To reduce transient fluctuations, a gesture transition was triggered only after the same prediction appeared in three consecutive time windows.

The robotic execution system comprised a 7-DoF KUKA LBR iiwa 14 R820 collaborative manipulator (KUKA, Germany) equipped with an RH56BFX-2R five-fingered robotic hand (Beijing Inspire Robots Technology, China), featuring 12 joints and 6 degrees of freedom. Communication between the MI-DHMI and the robotic platform was implemented via TCP and ROS2, and an optical motion capture system was used to record human hand poses.

\section{Results}
\label{sec:IV}
\subsection{Gesture Decoding}

\begin{figure}[!t]  
\centering
\includegraphics[width=\columnwidth]{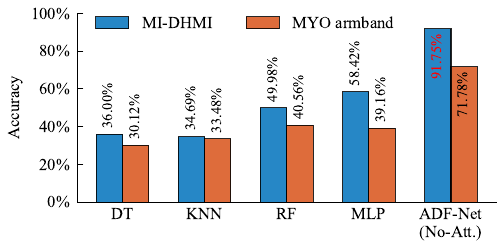}
\caption{Comparison of single-modality sEMG gesture decoding performance using our ADF-Net without attention and classical machine learning models on the MI-DHMI and MYO armband.}
\label{fig_7}
\end{figure}

\begin{figure}[!t]  
\centering
\includegraphics[width=\columnwidth]{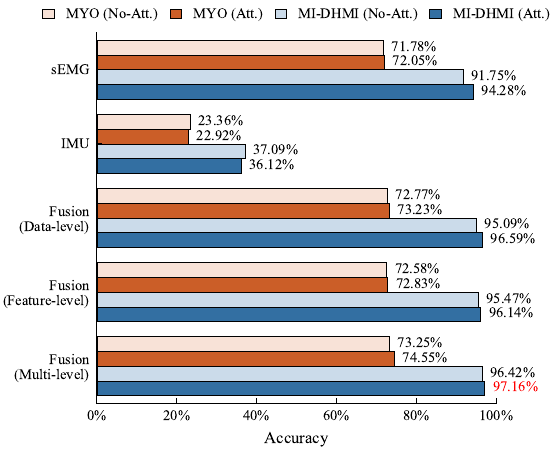}
\caption{Gesture decoding accuracy comparison under different fusion strategies and attention mechanisms, including single modality sEMG decoding, data-level fusion, feature-level fusion, and their combination, evaluated on the HI-DHMI and the MYO armband.}
\label{fig_8}
\end{figure}

\begin{figure*}[!t]
\centering
\includegraphics[width=\textwidth]{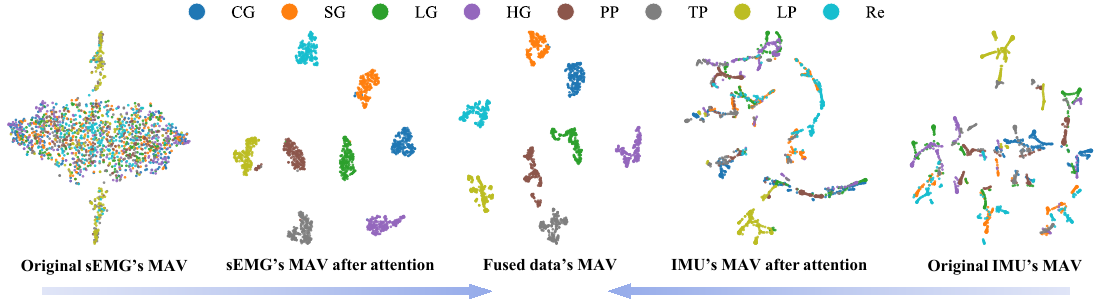}
\caption{Visualization of feature distributions illustrating the effects of
modality selection, attention mechanisms, and multimodal fusion on feature
separability.}
\label{fig_9}
\end{figure*}

\begin{figure}[!t]  
\centering
\includegraphics[width=\columnwidth]{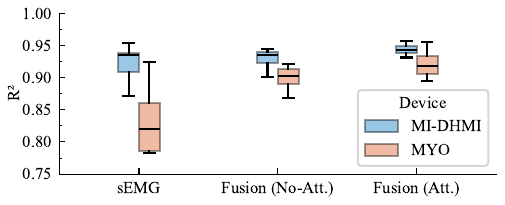}
\caption{Comparison of force regression performance across different devices
and modalities, including single modality sEMG decoding and multimodal fusion of sEMG and IMU signals with and without the attention mechanism.}
\label{fig_10}
\end{figure}

As shown in Fig.~\ref{fig_7}, classical machine learning models relying solely on sEMG achieved limited performance for both MI-DHMI and the MYO armband, with average accuracies ranging from 30.16\% to 58.42\%. In contrast, the proposed ADF-Net substantially improved sEMG-based gesture decoding, reaching an accuracy of 91.75\%, which further increased to 94.28\% with the attention mechanism (Fig.~\ref{fig_8}), demonstrating its robustness under unconstrained wrist and forearm motion. Incorporating IMU quaternion data led to additional performance gains. Data-level fusion achieved accuracies of 95.09\% (96.59\% with attention), while feature-level fusion reached 95.47\% (96.14\% with attention). The highest accuracy was obtained by combining both fusion strategies, achieving 96.42\% without attention and 97.16\% with attention.

To analyze the effects of multimodal fusion and the attention module in ADF-Net, feature distributions were visualized using t-distributed stochastic neighbor embedding (t-SNE) on five types of feature vectors. For each sEMG or IMU window, temporal averaging produced raw single-modality vectors, while attention-enhanced features were obtained by temporally averaging the pooled representations before the final fully connected layers. These five feature types include raw single-modality data, attention-enhanced single-modality features, and attention-enhanced multimodal fusion features, enabling a systematic comparison. As shown in Fig.~\ref{fig_9}, single-modality sEMG or IMU features exhibited limited separability due to unconstrained wrist motion and similar grasping patterns. The attention mechanism improved clustering by emphasizing informative channels, although some sEMG samples remained ambiguous and IMU features alone were still weakly discriminative. In contrast, attention-based multimodal fusion produced well-separated and compact clusters, demonstrating that IMU signals effectively complement sEMG for distinguishing challenging gestures.

\subsection{Force Regression}
In the force regression task, IMU-only decoding showed negligible correlation, as IMU signals mainly capture unconstrained wrist motion rather than force-related activity. Therefore, we evaluated single-modality sEMG and multimodal fusion strategies. The decoding performance across devices and modalities is summarized in Fig.~\ref{fig_10}. For single-modality sEMG, MI-DHMI consistently achieved higher $R^2$ values (0.87–0.95) than the MYO armband (0.78–0.92), confirming sEMG as the primary modality for grasp force decoding. When multimodal fusion was performed without attention, the inclusion of IMU signals improved decoding accuracy for both devices, yielding $R^2$ values of 0.90–0.94 for MI-DHMI and 0.87–0.92 for MYO. The highest performance was obtained with attention-based multimodal fusion, where $R^2$ values reached 0.94–0.95 for MI-DHMI and 0.90–0.95 for MYO, indicating that attention further enhances multimodal force decoding.

\subsection{Hand Pose Estimation}

\begin{figure}[!t]  
\centering
\includegraphics[width=\columnwidth]{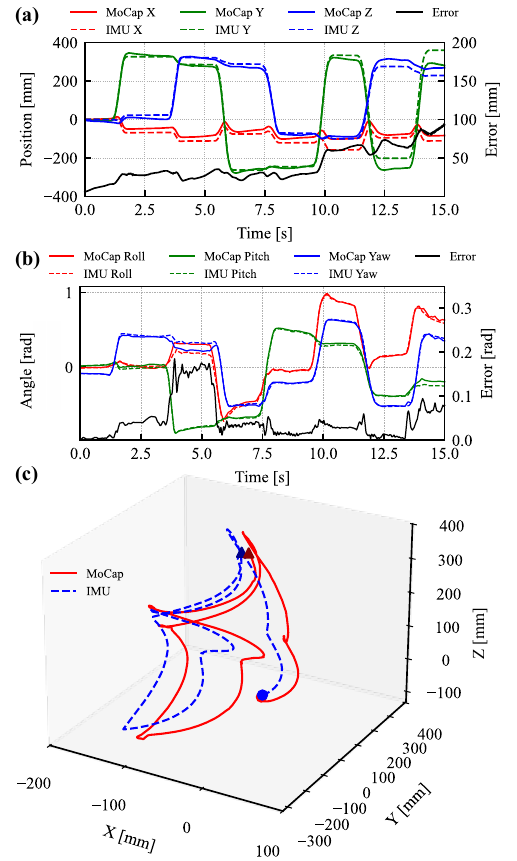}
\caption{A comparison chart of the real hand poses obtained from the motion capture system and the predicted hand poses calculated based on the hand-mounted IMU. MoCap: motion capture system.}
\label{fig_11}
\end{figure}

During hand pose estimation, we observed that directly integrating the acceleration leads to trajectories that are highly susceptible to noise. By applying the SZPR method for hand pose reconstruction, the decoded hand poses are shown in Fig.~\ref{fig_11}. Position decoding exhibits a drift over time, as illustrated in Fig.~\ref{fig_11}(a). In contrast, the decoding error for orientation is relatively small and is compensated during movement, as shown in Fig.~\ref{fig_11}(b). The reconstructed trajectories compared with the ground-truth trajectories are presented in Fig.~\ref{fig_11}(c).

\subsection{Teleoperation Experiments}

\begin{figure}[!t]  
\centering
\includegraphics[width=\columnwidth]{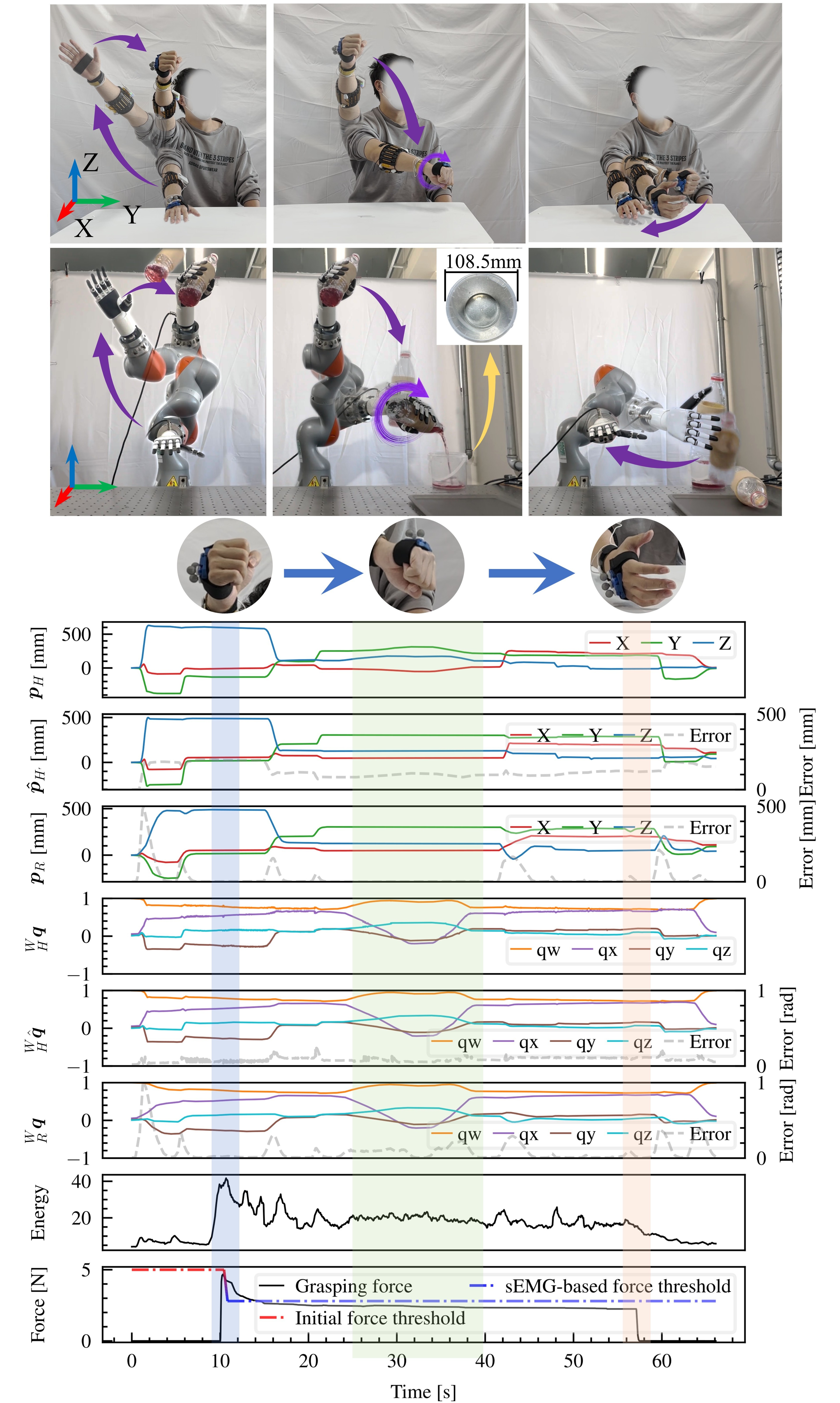}
\caption{Illustration of the teleoperation experiment process and corresponding signals. Each row of subplots presents: 
(1) ground-truth human hand positions captured by Vicon, 
(2) human hand positions estimated from the hand-mounted IMU (with errors relative to Vicon measurements), 
(3) robotic hand positions (with errors relative to IMU-estimated positions), 
(4) human hand orientations from Vicon, 
(5) human hand orientations estimated from the IMU (errors relative to Vicon orientations), 
(6) robotic hand orientations (errors relative to IMU-estimated orientations), 
(7) 64-channel sEMG energy map, and 
(8) comparison of sEMG-based predicted grasping force and actual robotic grasping force.}
\label{fig_12}
\end{figure}

\begin{figure}[!t]  
\centering
\includegraphics[width=\columnwidth]{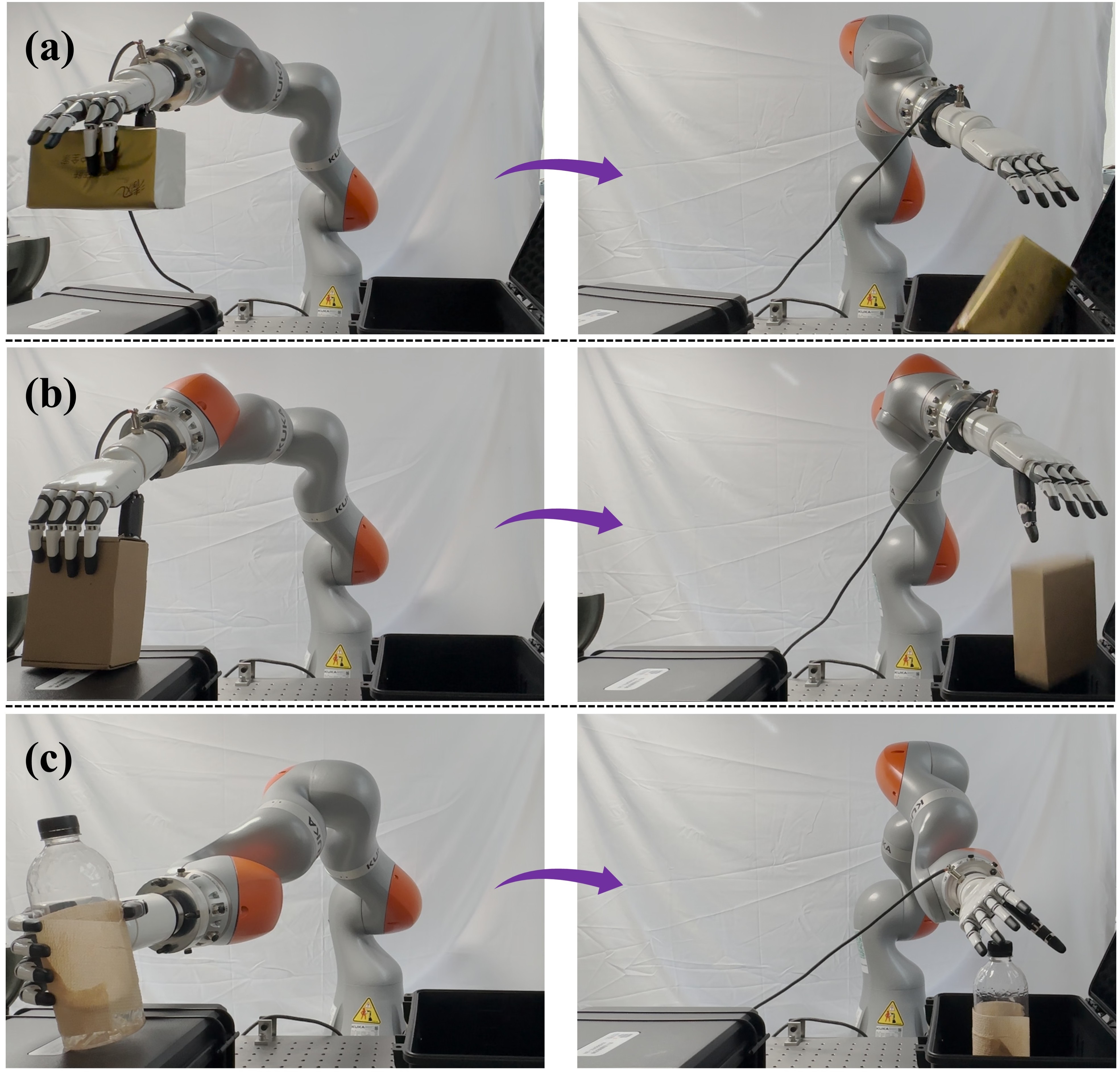}
\caption{Illustration of grasping and releasing actions for three representative gestures during the teleoperation task. The manipulator's end-effector executes the predicted action and places the object at the designated location.}
\label{fig_13}
\end{figure}

The first teleoperation task involved grasping and pouring water. To evaluate performance, hand and robotic positions and orientations, sEMG signals, and grasping force were recorded, as shown in Fig.~\ref{fig_12}. sEMG signals were characterized by energy, computed as the root mean square of each channel within a time window, averaged across all channels. For a window $\boldsymbol{S} \in \mathbb{R}^{T \times C_s}$, the energy is given by:
\begin{equation}
Energy = \frac{1}{C_s} \sum_{c=1}^{C_s} \sqrt{\frac{1}{T} \sum_{t=1}^{T} S_{t,c}^2 }.
\label{eq:emg_energy}
\end{equation}

During the water-pouring task, trajectories estimated from the hand-mounted IMU generally followed the ground-truth positions from the motion capture system, with deviations partially compensated by human-in-the-loop adjustments. Robotic hand trajectories showed minor deviations, mainly during arm motion due to joint‑velocity limits, and after temporal alignment, the robotic arm completed the task with near-millimeter positional accuracy. Orientation tracking indicated that IMU estimates reasonably reflected the motion capture measurements and robotic end-effector orientations.

The 64-channel sEMG energy map reflected muscle activation and temporal fluctuations. Despite wrist and forearm motion, ADF-Net maintained stable gesture recognition. During grasping, no unintended hand opening or object slippage occurred, and the sEMG-based predicted force effectively regulated the robotic hand to prevent excessive or insufficient force, ensuring stable manipulation. During the maintenance phase, the operator did not need to maintain a specific force, reducing physical workload.

For the online teleoperation task evaluating multiple grasping gestures, each gesture class was tested through ten repeated grasping trials to assess the robustness of action-level teleoperation, as shown in Fig.~\ref{fig_13}. The corresponding success rates are summarized in Table~\ref{tab:gesture_success}. The LP gesture exhibited comparatively lower success rates, primarily due to their higher requirements on positional accuracy during grasp execution. In contrast, gestures with less stringent positional precision requirements achieved higher success rates.

\begin{table}[!t]
\centering
\caption{Success rates of online grasping tests}
\label{tab:gesture_success}
\resizebox{\columnwidth}{!}{%
\begin{tabular}{c c c c c c c}
\hline
\textbf{CG} & \textbf{SG} & \textbf{LG} & \textbf{HG} & \textbf{PP} & \textbf{TP} & \textbf{LP} \\
\hline
10 / 10 & 8 / 10 & 9 / 10 & 9 / 10 & 8 / 10 & 8 / 10 & 7 / 10 \\
\hline
\end{tabular}%
}
\end{table}

\section{Discussion}
\label{sec:V}
In summary, we present a novel wearable MI-DHMI capable of reliably decoding multiple hand intentions under unconstrained wrist and forearm motion. The lightweight, non-optical interface and its decoding framework were experimentally validated. As shown in Table~\ref{tab:comparison_related_work}, this system uniquely enables simultaneous estimation of gesture, force, and pose. To the best of our knowledge, this is the first fully wearable multimodal HMI for ubiquitous teleoperated grasping capable of decoding multiple hand intentions under unconstrained wrist and forearm motion.

Multimodal fusion enhances decoding robustness by integrating complementary wrist kinematics. IMU signals provide motion cues absent from sEMG, and joint modeling of muscle activity and wrist kinematics reduces rotation-related interference, improving feature separability for ambiguous sEMG samples. Both data-level and feature-level fusion contribute to this effect. A similar benefit is observed in grasp force decoding, where IMU signals help mitigate motion-induced sEMG variability while preserving force-related information. Overall, attention-enhanced multimodal fusion effectively improves robustness for both gesture classification and force regression under dynamic conditions.

In addition, MI-DHMI exhibits clear advantages as a fully wearable interface. The system operates without external cameras, markers, or environmental infrastructure, enabling reliable use under unconstrained and dynamic conditions. Compared with commercially available sEMG-based HMIs, MI-DHMI provides higher signal throughput and fidelity, as well as integrated kinematic sensing, which together support more robust and informative hand intention decoding. These advantages make MI-DHMI better suited for practical teleoperation scenarios requiring mobility, portability, and consistent performance.

Although the proposed framework achieved promising performance under unconstrained wrist motion, its cross-subject generalization remains to be explored. Future work will investigate subject-independent and adaptive training strategies to improve scalability. In addition, while IMU signals were used to provide wrist kinematics, incorporating additional forearm IMUs and adaptive fusion schemes may further improve the separation of motion- and task-related muscle activity. Finally, beyond decoding accuracy, future studies will focus on deployment on edge or embedded platforms, with systematic evaluation of latency, power consumption, and long-term stability in real-world teleoperation scenarios.

\begin{table}[!t]
\centering
\caption{Comparison with Related Works}
\label{tab:comparison_related_work}
\renewcommand{\arraystretch}{1.15}
\setlength{\tabcolsep}{4pt}
\resizebox{\columnwidth}{!}{
\begin{tabular}{l c c c c c c c}
\hline
\textbf{Ref.} &
\textbf{Modality} &
\textbf{Wearable} &
\textbf{Gesture} &
\textbf{Force} &
\textbf{Pose} &
\makecell{\textbf{Free}\\\textbf{Upper-Limb}} &
\makecell{\textbf{Free}\\\textbf{Wrist}} \\
\hline
\cite{anipulator-based} & \makecell{Mechanical motion\\\ + Force/Torque}\  & $\times$ & $\times$ & $\checkmark$ & $\checkmark$ & $\checkmark$ & $\checkmark$ \\
\cite{li2025six} & Optical + Data glove & $\circ$ & $\checkmark$ & $\times$ & $\checkmark$ & $\checkmark$ & $\checkmark$ \\
\cite{li2020mobile} & Optical + IMU & $\checkmark$ & $\checkmark$ & $\times$ & $\checkmark$ & $\checkmark$ & $\checkmark$ \\
\cite{sEMG_gesture_force} & sEMG & $\checkmark$ & $\checkmark$ & $\checkmark$ & $\times$ & $\times$ & $\times$ \\
\cite{sEMG_pFMG_8834} & sEMG + pFMG & $\checkmark$ & $\checkmark$ & $\times$ & $\times$ & $\circ$ & $\times$ \\
\cite{lv2025egohand} & Radar + IMU & $\checkmark$ & $\checkmark$ & $\times$ & $\checkmark$ & $\checkmark$ & $\checkmark$ \\
\hline
\makecell{\textbf{Our}\\\textbf{work}} & sEMG + IMU & $\checkmark$ & $\checkmark$ & $\checkmark$ & $\checkmark$ & $\checkmark$ & $\checkmark$ \\
\hline
\end{tabular}}
\addvspace{1mm}
\footnotesize{$\checkmark$: full; $\circ$: partial; $\times$: unavailable.}
\end{table}

\section{Conclusion}
\label{sec:VI}
In this work, we developed a fully wearable MI-DHMI capable of simultaneously decoding hand pose, gestures, and grasping force under unconstrained wrist and forearm motion, enabling its use in ubiquitous teleoperated grasping tasks. The interface consists of an armband integrating high-throughput sEMG sensors and an IMU, together with a hand-mounted IMU, supported by a unified framework for simultaneous intentions decoding. By fusing wrist angle data from the IMUs with sEMG signals capturing muscle activations, the interface robustly decodes hand intentions even under unconstrained wrist and forearm motion. Ablation studies and comparative experiments demonstrate that the developed armband outperforms commercial sEMG-based devices, and also highlight the effectiveness of the hardware design, multimodal fusion, and attention modules. Teleoperation tests further demonstrate practical feasibility, with operators successfully performing water pouring and multi-gesture object manipulation. Together, these results establish the MI-DHMI as a practical, multimodal platform for ubiquitous teleoperation and a promising foundation for advancing human–machine interaction research.

\bibliographystyle{bibtex/IEEEtran}
\bibliography{bibtex/my_config, bibtex/references_pure, bibtex/IEEEabrv}

\vfill

\end{document}